\documentclass[12pt]{article}
\usepackage{graphicx}
\usepackage{setspace}
\usepackage{amsmath}
\usepackage{amssymb}
\usepackage{fullpage}
\usepackage{xcolor}
\usepackage{verbatim}
\usepackage{wasysym}
\usepackage{listings}
\usepackage{float}
\usepackage[T1]{fontenc}
\usepackage{geometry}
\usepackage{url}
\usepackage{fancyhdr}
\begin{document}
\title{Measurement Models For Sailboats Price vs. Features And Regional Areas}
\author{Jiaqi Weng, Chunlin Feng, Yihan Shao}
\date{March, 2023}
\maketitle
\noindent
Sailboats are a timeless symbol of adventure and freedom on the open sea, but choosing the right boat can be a daunting task. In this research paper, we explore the relationship between the technical specifications of sailboats and their prices, as well as the impact of regional factors on sailboat pricing. Using a variety of machine learning models, we developed a predictive model that can help potential buyers make informed decisions. \\
Our dataset included sailboats' technical specifications, such as length, beam, draft, displacement, sail area, and waterline. After testing multiple machine learning models, including multiple variable linear regression and ADADELTA models, we found that the gradient descent model outperformed them all, with the lowest mean squared error (MSE) and mean absolute error (MAE).\\
After conducting a correlation analysis, we discovered several interesting findings about the relationship between sailboats' technical specifications and their prices. We found that monohulled boats tend to be less expensive than catamaran boats, and that sailboats with longer length, beam, waterline, displacement, and sail area tend to have higher listing prices, while boats with lower draft tend to have higher listing prices. \\
In addition to exploring the relationship between technical specifications and prices, we also examined the impact of regional factors on sailboat prices. After testing our initial hypothesis that a country's GDP might be a good indicator of its sailboat market, we found no clear relationship between GDP and sailboat listing prices. Using a dummy variable method to assign numerical values to regions, we trained our model to analyze the relationship between region and listing price. Our findings revealed that the United States had the highest average listing price for sailboats overall, followed by Europe, Hong Kong, and the Caribbean Sea. \\
In the end, we used fifty percent cross validation method to test our models, and both models were passed the test successfully with no more then ten present difference when crossing testing groups and modelling groups. \\
Overall, our research provides valuable insights into the complex interplay between sailboats' technical specifications, regional factors, and their prices. By using our predictive model to evaluate potential sailboat purchases, buyers can make more informed decisions and ensure they find the boat that best fits their needs and budget. This research also sheds light on the important role of machine learning in the boating industry, demonstrating how it can be used to analyze large amounts of data and make accurate predictions about complex systems.
\newpage

\tableofcontents
\newpage
\section{Introduction}
\subsection{Background}

Sailboats, also known as sailing yachts, are a type of watercraft that use wind as their primary source of propulsion. They come in various sizes and shapes, ranging from small dinghies to massive ocean-going vessels, and are designed for different purposes such as cruising, racing, or exploring.\\
One important aspect of sailboats is their technical specifications. These include factors such as the length overall, the beam (width), draft (the distance from the waterline to the bottom of the boat), displacement (the weight of the boat), sail area (the total area of the sails), etc. These technical specifications can greatly affect the performance and handling of the boat, as well as its listing price.\\
For example, a sailboat with a longer length generally has more interior space and can handle larger sail areas, making it suitable for longer cruises or racing and more expensive. A wider beam can provide greater stability and more room for passengers. A sailboat's displacement can also impact its performance, with lighter boats being faster and more maneuverable, but heavier boats generally offering better stability and comfort in rough waters. 
\subsection{Problem Restatement}
The objective of this project is to develop a mathematical model that explains the listing price of each sailboat variant provided in the spreadsheet. The model should include any relevant predictors, such as the boat's characteristics, location, and other economic data by year and region [2]. The sources of data used should be identified and described, and the precision of the estimates for each sailboat variant's price should be discussed. \\
Furthermore, this project aims to analyze the effect of the sailboat's geographic region on its listing price. It will be determined whether any regional effects are consistent across all sailboat variants and whether such effects are statistically and practically significant.\\
Another objective of this project is to evaluate how the modeling of different geographic regions can be useful in the Hong Kong (SAR) market. A subset of sailboats, including monohulls and catamarans, will be selected from the provided spreadsheet, and comparable listing price data for that subset will be obtained from the Hong Kong (SAR) market. The regional effect of Hong Kong (SAR) on each sailboat variant's price will be modeled to determine whether there is a regional effect and, if so, whether it is consistent across both monohull and catamaran sailboats.\\
Finally, any other interesting and informative inferences or conclusions drawn from the data will be identified and discussed. A one- to two-page report will be prepared for the Hong Kong (SAR) sailboat broker, which will include well-chosen graphics to help the broker understand the conclusions.
\subsection{Assumptions}
\begin{itemize}
\item We assume all information in the given table are correct.
\item We assume no external influential factors in the society are present to influence the price, such as economic depression or war.
\item We assume the price of sailboats are not influenced by the price of other sailboats.
\item We assume that the predictor variables are not highly correlated to one another.
\end{itemize}
\subsection{Definitions and Abbreviations}
\begin{itemize}
\item MSE: mean squared error
\item MAE: mean absolute error
\item ADADELTA: an extension algorithm to ADAGRAD, or adpative gradient algorithm.
\end{itemize}
\section{Data Collection And Cleanings}
In the present study, data cleaning was performed using a web crawler tool, named Octoparse[4], to acquire essential data pertaining to original boats available on the official COMAP website. The website we were crawling from includes BoatTrader[11], Sailboatdata[12], and Yachtworld[13]. In addition to the pre-existing information encompassing making year, length, and variant, we incorporated five additional technological specifications, namely beam, draft, displacement, sail area, and waterline length. Furthermore, a categorical variable was generated to indicate the sailboat's hull type, classified as either a catamaran or monohull. 

In addition to the 2,347 data points available on the official website, a further 1,331 data points were randomly selected from the aforementioned websites, resulting in a total of 3,678 raw data points regarding these sailboats. Upon completion of data collection, initial data cleaning processing was performed, recognizing some data points had poorly preserved information due to web crawling limitations. As a result, objects with a high degree of data loss were removed through direct deletion, including twenty sailboats due to missing regional information and 147 sailboats due to missing technical information.

\section{The Models}
\subsection{Models For Listing Prices vs. Sailboats Technology Specifications}
\subsubsection{Correlation Analysis}
Prior to conducting further analyses, we performed a correlation check between the eight variables collected and their corresponding listing price. To illustrate these relationships, eight graphs were plotted, including a simple trending line [3]. The resulting plots are presented below.
\begin{center}
\includegraphics[scale = 0.3]{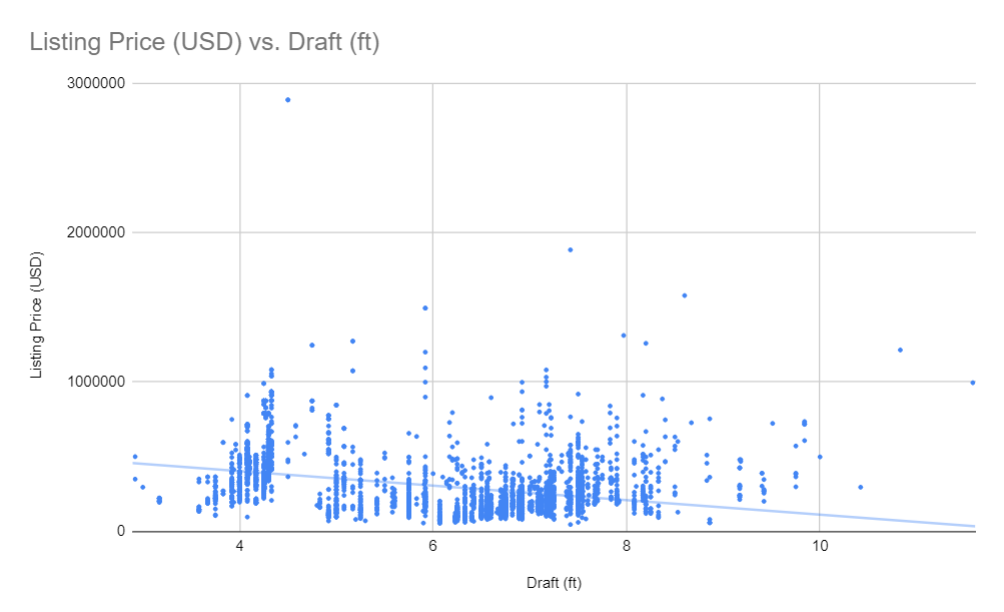}
\includegraphics[scale = 0.3]{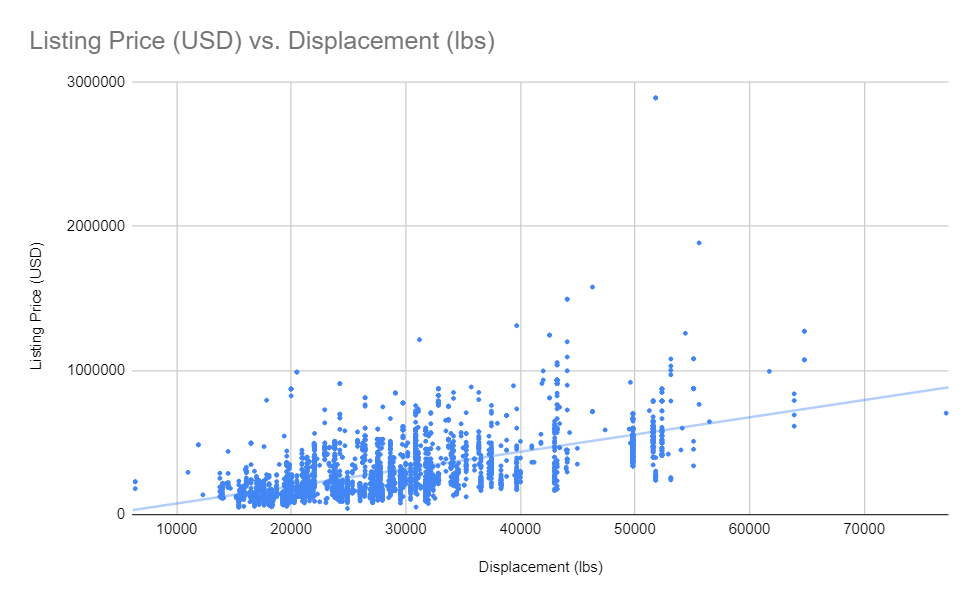} \\ 
Figure 1 \quad \quad \quad \quad \quad \quad \quad \quad \quad \quad \quad \quad \quad \quad \quad \quad  Figure 2 \\ 
\includegraphics[scale = 0.3]{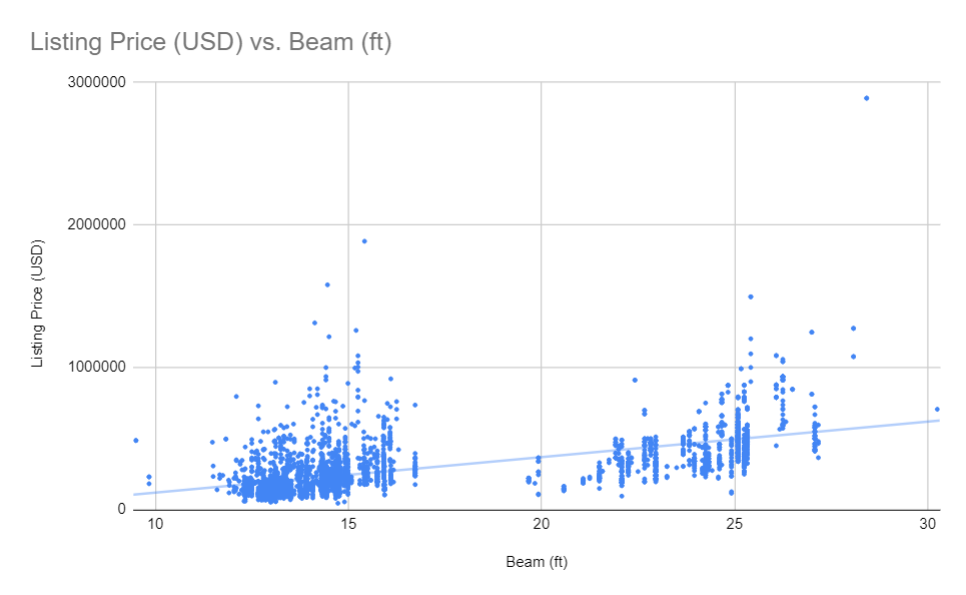}
\includegraphics[scale = 0.3]{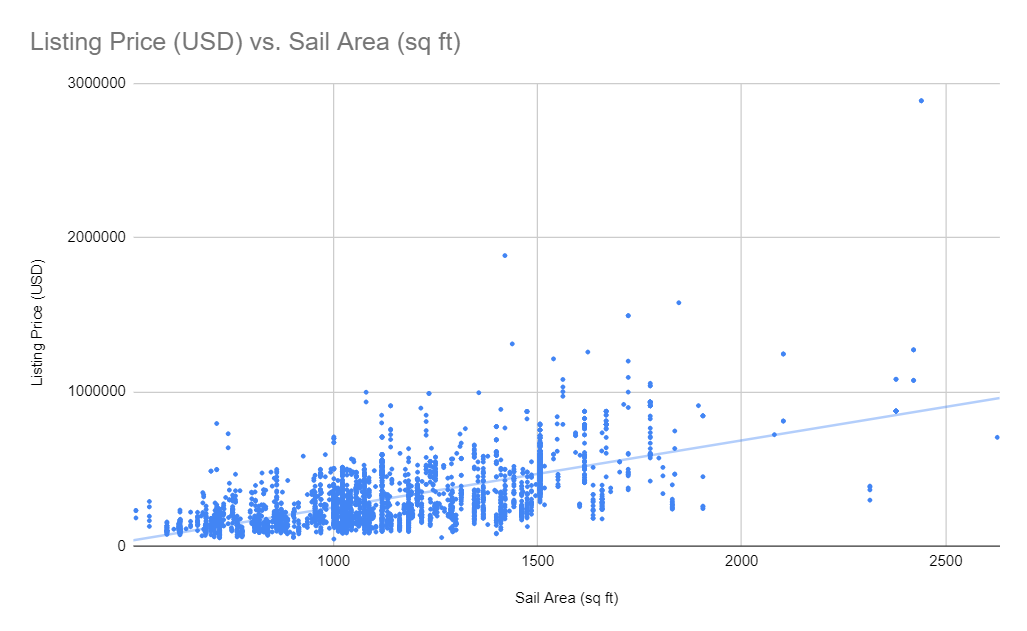} \\ 
Figure 3 \quad \quad \quad \quad \quad \quad \quad \quad \quad \quad \quad \quad \quad \quad \quad \quad  Figure 4 \\ 
\includegraphics[scale = 0.3]{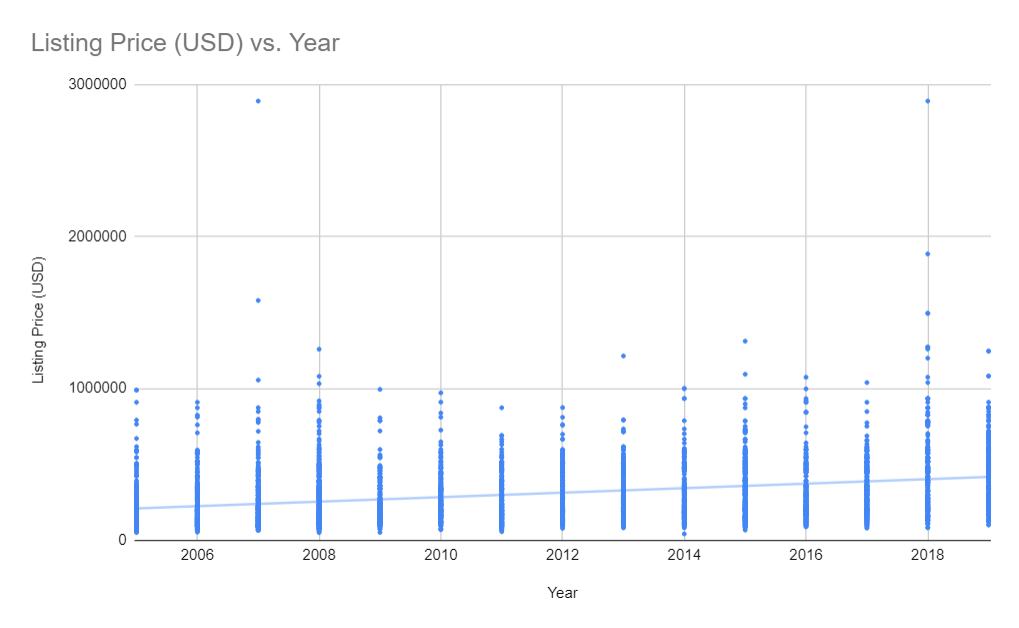}
\includegraphics[scale = 0.3]{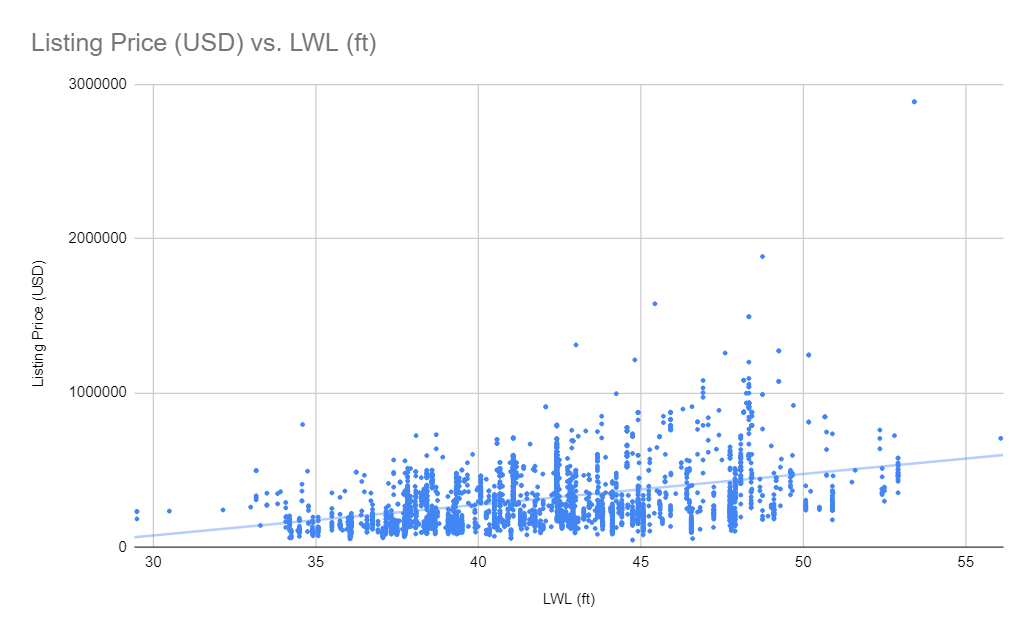} \\ 
Figure 5 \quad \quad \quad \quad \quad \quad \quad \quad \quad \quad \quad \quad \quad \quad \quad \quad  Figure 6 \\ 
\includegraphics[scale = 0.3]{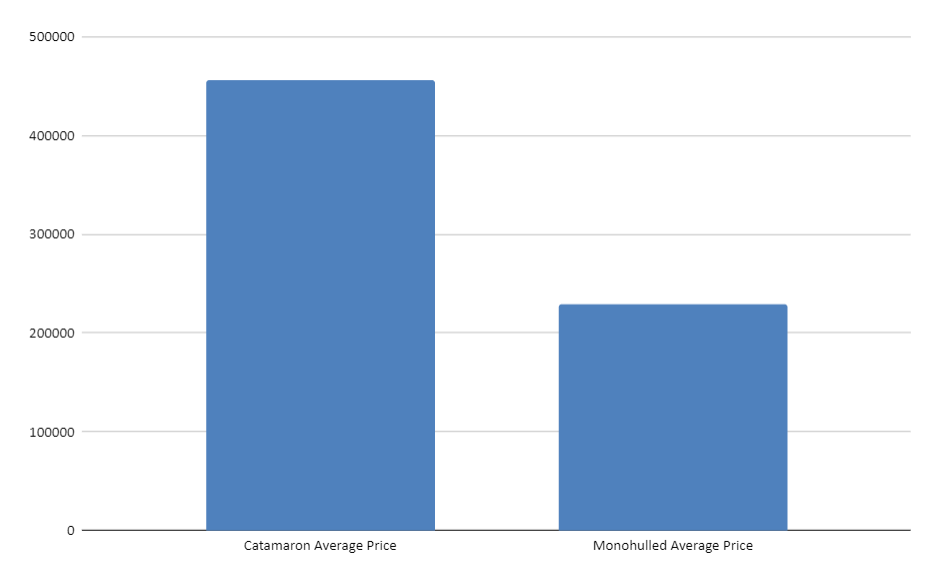}
\includegraphics[scale = 0.3]{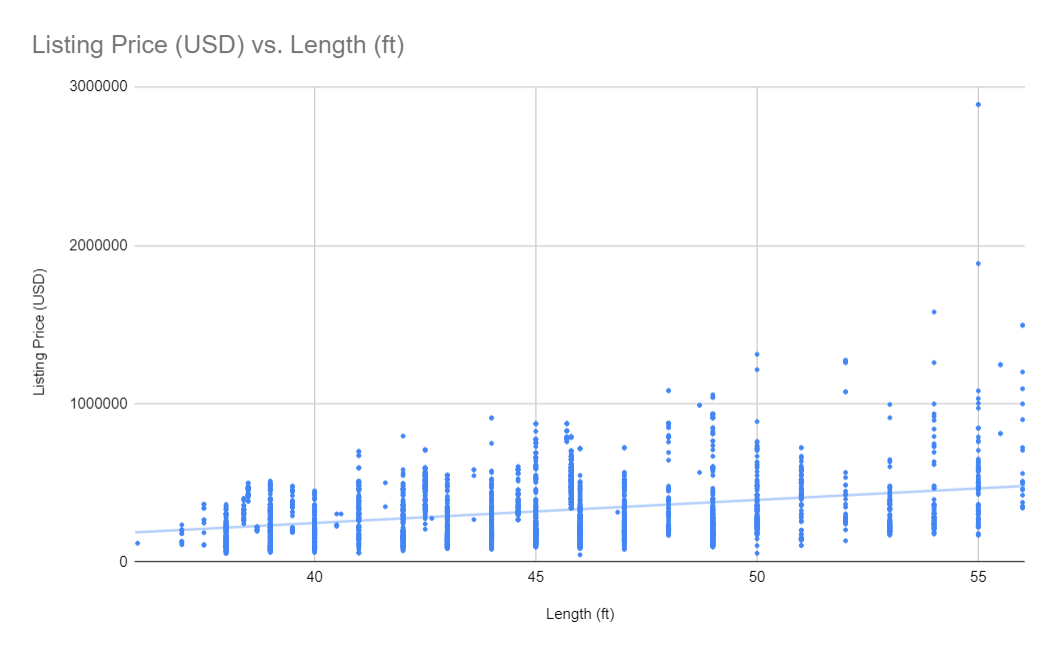} \\ 
Figure 7 \quad \quad \quad \quad \quad \quad \quad \quad \quad \quad \quad \quad \quad \quad \quad \quad  Figure 8
\end{center}
Based on the observation of the provided plots, it is evident that a strong positive correlation exists between displacement, beam, length, year, waterline, and sail area and their corresponding listing price [4]. Conversely, a negative correlation is observed in the relationship between draft and the listing price. Additionally, the bar graph indicates a marked difference in the average listing price between catamaran and monohulled boats, with the former exhibiting a significantly higher average price. Since all the factors are influencing the listing price to a certain degree, we will include all of them in our following model selection. 
\subsubsection{Model Selection}
To evaluate the efficacy of our models, a simple cross-validation approach will be employed [5]. This technique involves dividing our dataset into two equal halves, generating a model based on the first half, and subsequently testing the residual, MSE, and MAE on the second half. By comparing and contrasting the results obtained from each model, we will select the most suitable model for our analysis. Residual means the difference between the model's output with the actual value of the testing group. MSE stands for mean squared error and MAE stands for mean absolute error. They are the two commonly used evaluation metrics in regression tasks. MSE measures the average of the squared differences between the predicted values and the actual values in the test set [6]. It is calculated by taking the sum of squared residuals and dividing by the number of observations:

\begin{equation}
MSE = \frac{1}{n} \sum_{i=1}^{n} (y_i - \hat{y}_i)^2
\end{equation}

where $n$ is the number of observations, $y_i$ is the actual value of the $i^{th}$ observation, and $\hat{y}_i$ is the predicted value of the $i^{th}$ observation.

On the other hand, MAE measures the average of the absolute differences between the predicted values and the actual values in the test set. It is calculated by taking the sum of absolute residuals and dividing by the number of observations:

\begin{equation}
MAE = \frac{1}{n} \sum_{i=1}^{n} |y_i - \hat{y}_i|
\end{equation}

where $n$, $y_i$, and $\hat{y}_i$ are defined the same as in MSE. MSE and MAE provide valuable information about the accuracy of the prediction models. While MSE is more sensitive to outliers, MAE provides a more robust measurement of the error in the model. Therefore, by comparing both MSE and MAE of a given regression model, we can obtain a more complete aspect of analysis. 
\subsubsection{Multiple Variable Linear Regression Model}
Given the significant number of variables involved in the analysis, a multiple variable linear regression model was deemed appropriate at first. Accordingly, the corresponding calculations were performed using MATLAB, and the resulting formula is presented below:
\begin{equation}
P = -31439565a+24406b+15509c-17411d-14712e+18856f+g+144h+360535
\end{equation}
where $a-h$ correspond to the coefficients for length, year, waterline, beam, draft, displacement, sail area, and the catamaran/monohulled indicator, respectively. However, these coefficient is notably inconsistent with the findings of our prior correlation analysis [7]. Furthermore, the residual plots indicate an average residual of approximately $0.3\cdot 10^6 - 0.5\cdot 10^6$, which is a significant value, despite the considerable variance present in the dataset. The multiple variable linear regression model exhibits an MSE and MAE of $14363701268$ and $75393$, respectively. These findings suggest that the model is inadequate and requires further refinement.
\begin{figure}[htbp]  
\begin{center}
\includegraphics[scale = 0.33]{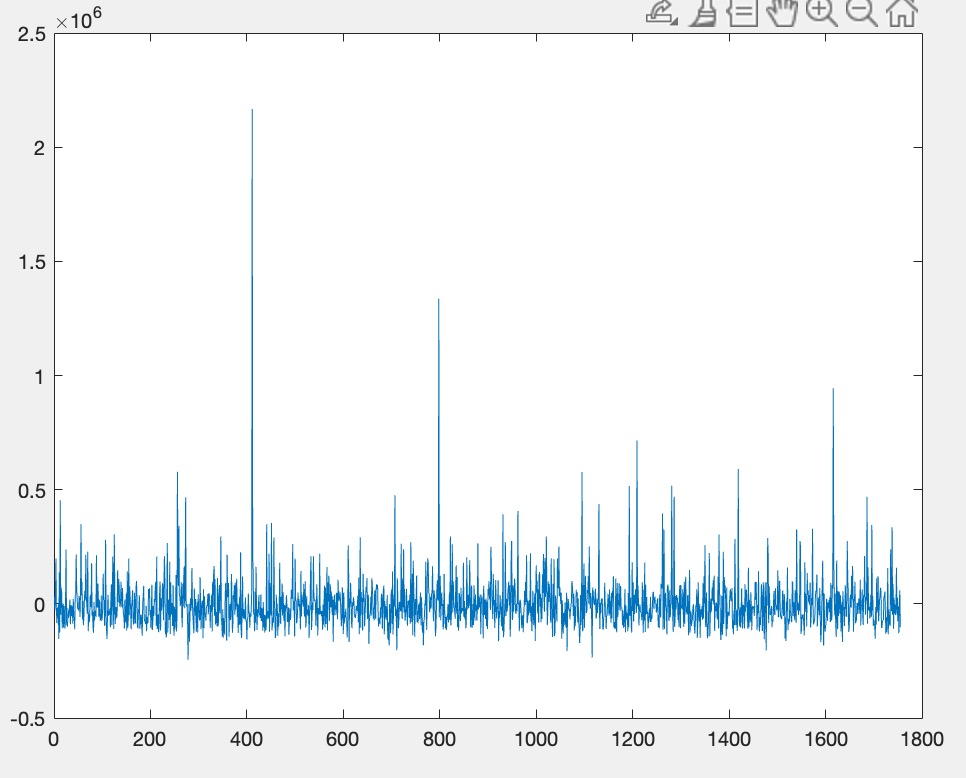}
\caption*{Figure 9: Positions trajectories and errors}
\end{center}  
\end{figure}
\subsubsection{ADADELTA Model}
Given that the conventional linear approximation methods did not yield satisfactory results, we have shifted our focus towards exploring machine learning techniques. As a result, we have chosen an optimization algorithm for training neural networks named ADADELTA. It maintains two sets of variables for each weight parameter: the average squared gradient and the average squared parameter update [8]. At the beginning of the training, these variables are initialized to zero. During each iteration of the training process, the gradients of the loss function with respect to the weights are calculated, and the average squared gradient is updated by taking the exponential moving average of the squares of the gradients. The update for the weight parameters is computed by dividing the root mean square of the parameter update by the root mean square of the gradients. The average squared parameter update is also updated by taking the exponential moving average of the squares of the parameter updates. Finally, the weights are updated by adding the computed update. ADADELTA adaptively adjusts the learning rate for each parameter based on a rolling average of the parameter updates and the gradients. Based on our cross validation method, the graph of residual is plotted below:
\begin{center}
\includegraphics[scale = 0.33]{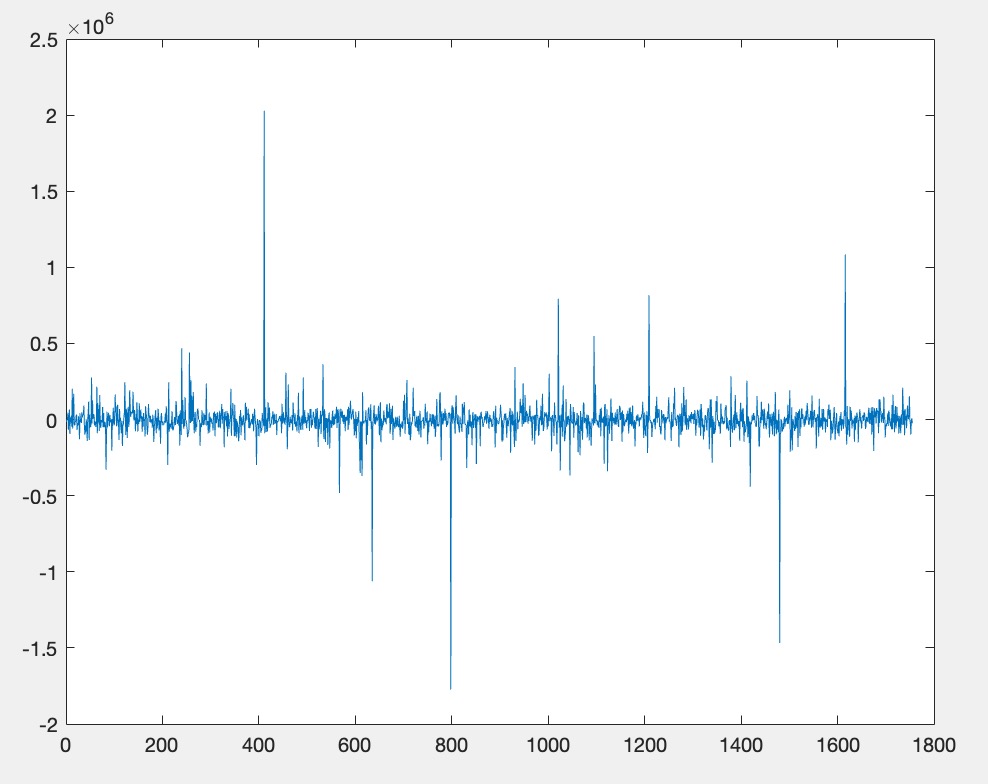} \\
Figure 10: Positions trajectories and errors
\end{center}  
Based on the observations of the graph, it is evident that the error residual remains high, with most values ranging between $-0.25 \cdot 10^6 - 0.25 \cdot 10^6$. Furthermore, the corresponding MSE and MAE values for our ADADELTA model are 13602449063 and 58032, respectively, representing only a minor improvement over the normal multiple variable linear regression model.
\subsubsection{Gradient Decent Regression}
Based on our trails on multiple different machine learning methods, we found that the gradient decent regression algorithm have the best fit on this dataset [9]. Gradient regression works by iteratively fitting a weak regression model to the negative gradient of the loss function with respect to the current model's predictions. The algorithm combines multiple weak models, which are simple regression models with low predictive power on their own, to create a stronger overall model. The weak models used in gradient regression are decision trees, a type of supervised learning algorithm. Gradient descent is used to optimize the loss function, and gradient boosting is used to combine the weak models. The final model is the combination of all the weak models, and its predictions are the sum of the predictions of each weak model. Gradient regression has several advantages over other regression algorithms, including its ability to handle a variety of data types and feature scales and its ability to capture complex nonlinear relationships between variables. However, it can be computationally expensive and may not perform as well on small data sets or data sets with highly correlated features. \\ \\ 
To implement our gradient regression algorithm, we followed the steps outlined below:
\begin{itemize}
    \item The model was initialized with a weak model, in this case, a decision tree with a small number of leaves.
    \item The negative gradient of the loss function with respect to the current model's predictions was computed.
    \item A new weak model was fitted to the negative gradient.
    \item The current model was combined with the new weak model by adding their predictions together.
    \item The iterative process from steps 2 to 4 was performed until a predetermined number of iterations were achieved, or until the loss function reached a minimum value indicating convergence.
\end{itemize}
The decision tree we used for our weaker model can be described as
\begin{equation}
f(x) = \sum_{i=1}^{L} y_i I(x \in R_i)
\end{equation}
where L is the number of leaf nodes, $y_i$ is the output value at the $i_{th}$ leaf node, $R_i$ is the region of the input space corresponding to the $i_{th}$ leaf node, and $I(x \in R_i)$ is an indicator function that takes the value 1 if x belongs to the region $R_i$ and 0 otherwise.\\
After initializing the base weak model for our dataset, we need to come up with a loss function in order to prepare for calculating the gradient later [10]. In our study, we have observed that the complexity of the dataset can lead to overfitting problems in machine learning models. Therefore, to mitigate this issue, we have adopted a regularization technique instead of using Mean Absolute Error (MAE) or Mean Squared Error (MSE) directly as our loss function. Specifically, we have employed the L2 regularization method, which is expressed as follows:
\begin{equation}
J(\theta) = MSE + \lambda \cdot sum(\theta^2)
\end{equation}
Here, $\lambda$ is the regularization parameter, $\theta$ is the vector of parameters derived from the weak regression models, and $\sum_{i=1}^{n}\theta_i^2$ represents the sum of the squares of the model parameters. This modified function is used as our actual loss function for further computation. \\
To minimize the loss function, we have employed the gradient descent method which is expressed as follows:
\begin{equation}
\theta = \theta - \alpha \cdot grad(J(\theta))
\end{equation}
Here, $\alpha$ represents the learning rate, $J(\theta)$ is the loss function, and $\text{grad}(J(\theta))$ represents the gradient of the loss function with respect to $\theta$. \\
We have employed this technique to our gradient boosting regression model. In this model, we have iteratively combined multiple weak models to create a stronger overall model. Each weak model is trained on the negative gradient of the loss function with respect to the current model's predictions. The formula for gradient boosting is as follows:
\begin{equation}
y_{pred} = y_{pred} + \alpha * f(x; \theta)
\end{equation}
Here, $y_{pred}$ is the current prediction, $f(x; \theta)$ represents the weak model (in this case, a decision tree), and $x$ is the input data.\\
We have used the aforementioned formulas to develop our gradient descent regression model. Our residual plot is presented in Figure 1. It is evident that our model exhibits the lowest residual among the three models tested. We have also calculated the MSE and MAE of our model using cross-validation methods, which were found to be 13602449063 and 58032, respectively. This is the lowest value among all the models tested, indicating the effectiveness of our proposed approach.

\begin{center}
    \includegraphics[scale = 0.4]{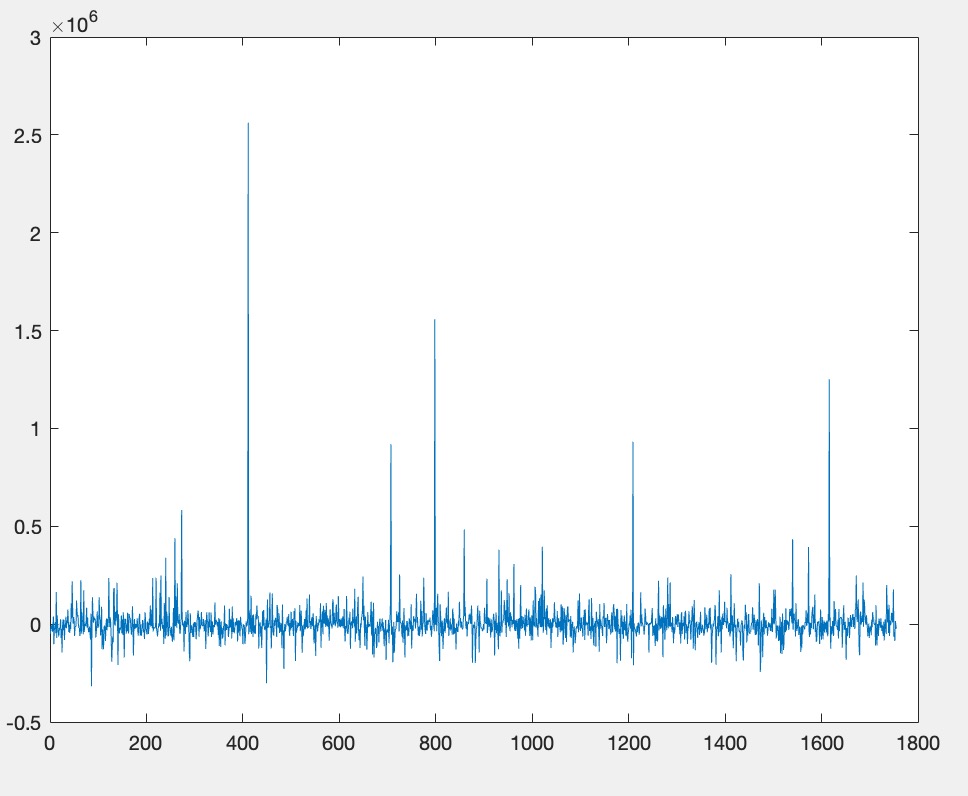}\\
    Figure 11: Residual plot for our gradient descent regression model.
\end{center}

\subsection{Model for Listing Price vs. Country \& Region}
\subsubsection{Model Analysis}
In addressing this research question, our team initially postulated that utilizing the Gross Domestic Product (GDP) and GDP per capita as key indicators would allow for a comprehensive evaluation of the impact of regional economic factors on sailboat pricing. Based on this premise, we formulated the hypothesis that a positive correlation exists between a region's GDP per capita and the average listing price of sailboats within that region. Therefore, we researched about the GDP and GDP per capita pertaining to the given country and year, which were added manually to the last two columns of our original data table. \\
In Section 3.1.5, a comprehensive comparison was conducted between the gradient descent regression model, ordinary multiple variable regression model, and the ADADELTA model. Based on the robust evidence presented, it was determined that the gradient descent regression model consistently outperforms the other two models. As a result, the gradient descent regression model was selected for subsequent analysis to investigate the influence of regional factors on listing prices. \\
In accordance with the methodology outlined in Section 3, an initial step in this analysis involved conducting a preliminary correlation analysis. This procedure was employed to provide an approximate understanding of the relationship between regional factors and listing prices. \\
\begin{center}
    \includegraphics[scale = 0.19]{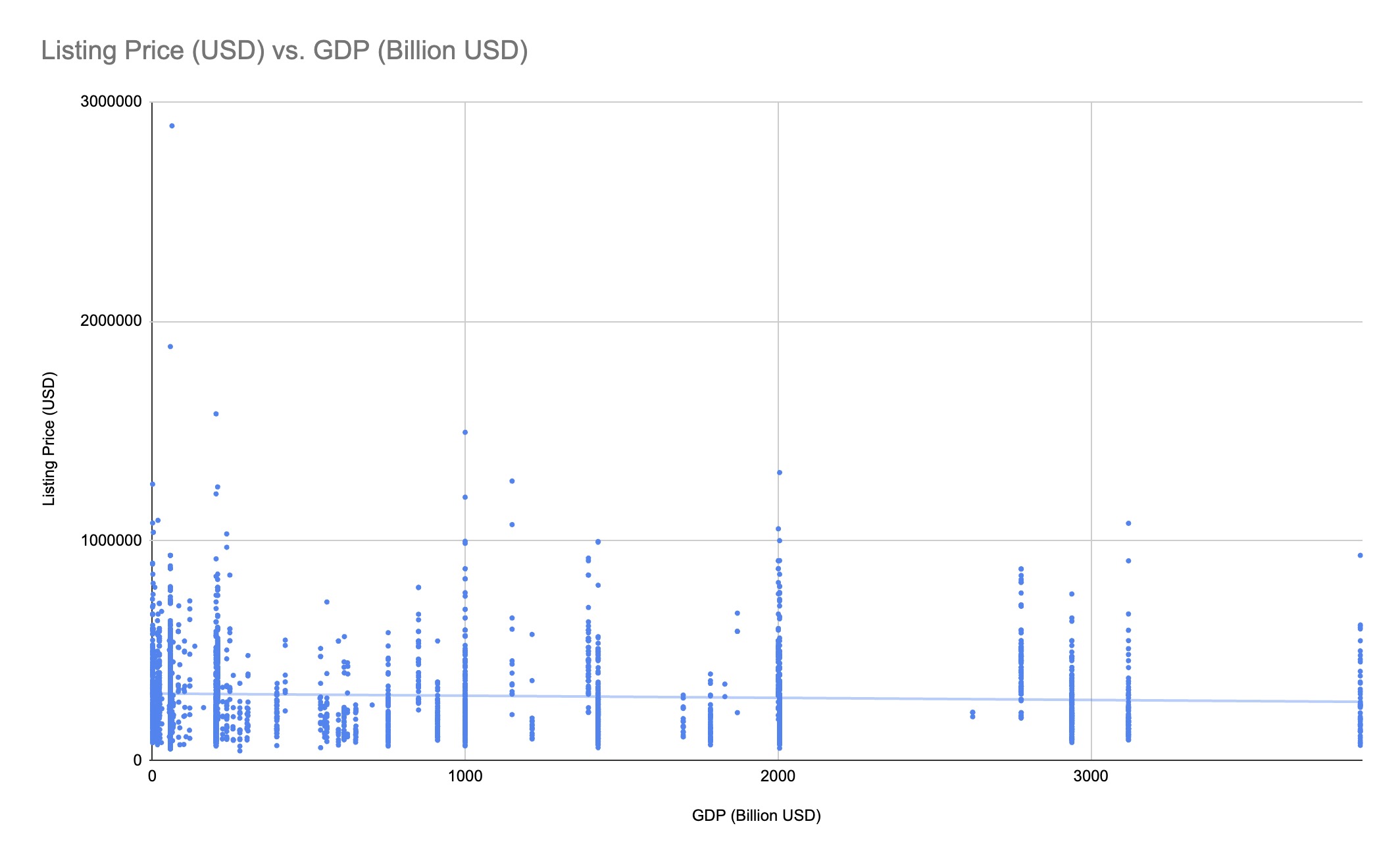}\\
    Figure 12 \\
    \includegraphics[scale = 0.18]{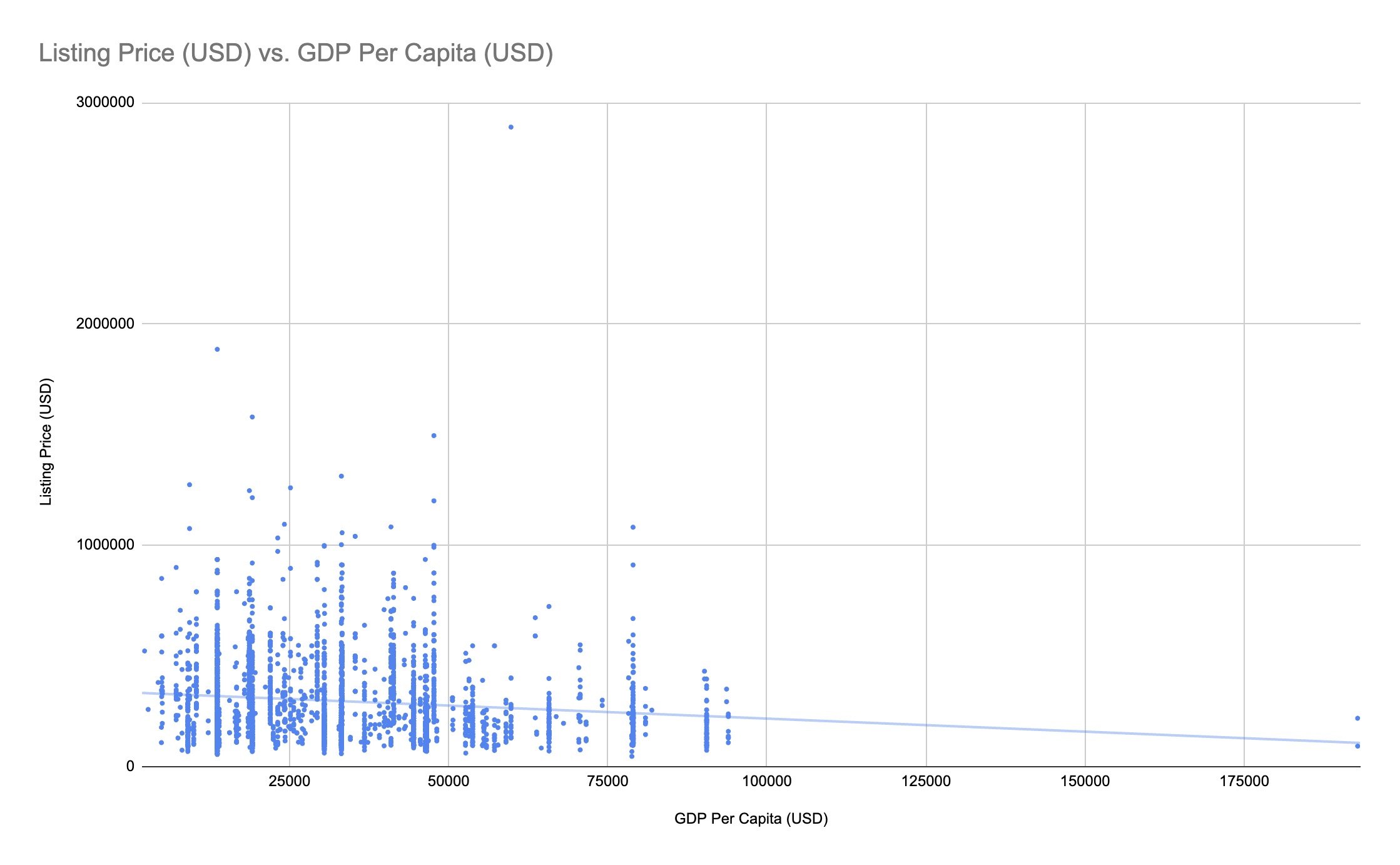}\\
    Figure 13 \\ 
\end{center}

Contrary to expectations, the two diagrams reveal a negligible correlation between GDP and GDP per capita in relation to the listing price of boats. This discrepancy may be attributed to an insufficient sample size. For instance, in Morocco, the country with the highest GDP per capita, only two sailboats are listed for sale. Thus, we proposed an alternative method to further investigate the impact of region on the listing price. In our study, we encountered a unique challenge with this region factor variable, which is a categorical variable represented by words rather than numerical values. As such, it cannot be directly incorporated into our regression model. To address this issue and ensure that this variable can be considered in our analysis, we implemented the dummy variable method. Specifically, we assigned a unique code to each region, and then incorporated this virtual variable as an input feature into our model. For example, we assigned the code 100 to the Caribbean region, 010 to Europe, and 001 to the USA. This approach allowed us to effectively incorporate the region factor into our model and obtain meaningful insights. 

\subsubsection{Linear Regression Model}
In Section 3, we conducted a comparative analysis of the performance of the gradient descent regression model, multiple variable linear regression model, and ADADELTA model. The results indicated that the gradient descent regression model outperformed the other two models. However, for the specific purpose of examining the relationship between the regional factor and the listing price, we employed the multiple variable linear regression model. The results of the modified model are presented in Figure below:\\
\begin{center}
    \includegraphics[scale = 0.37]{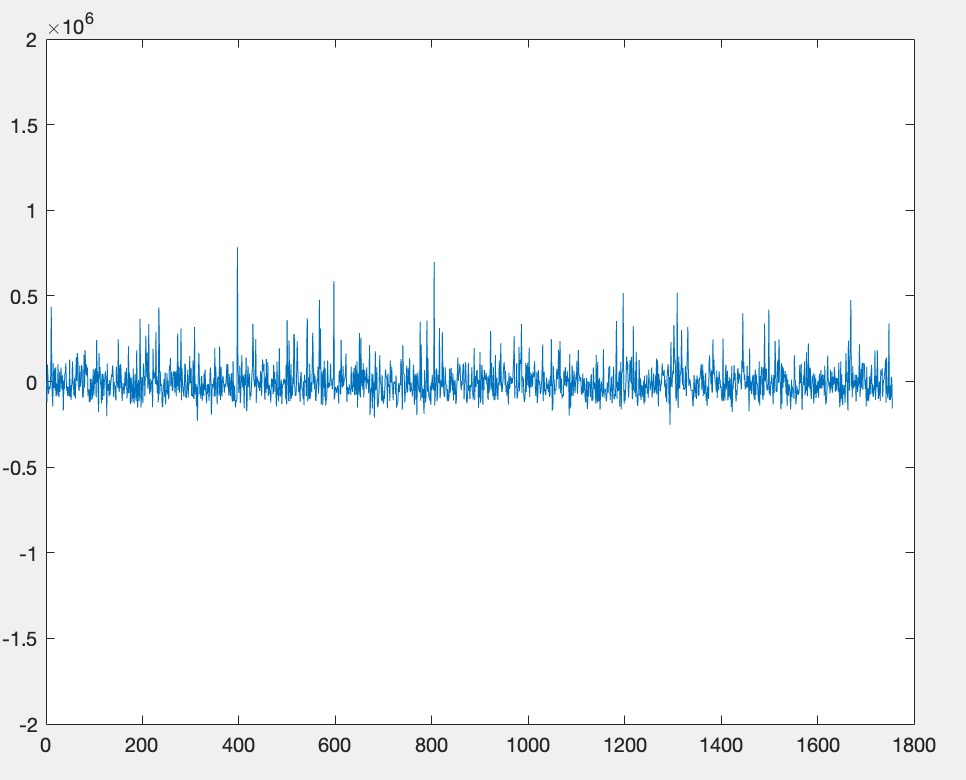}\\
Figure 14
\end{center}
It is noteworthy that the residuals are within an acceptable range, as evidenced by the low MSE and MAE values of 9964296633 and 52209, respectively. These values are significantly lower than those obtained in the previous analysis, which did not include the regional factor. Moreover, the positive coefficient for the region variable indicates a clear positive relationship among the three different regions. Specifically, using the Caribbean region as a base factor, the Europe and USA regions have coefficients of 17809.42 and 117553.40, respectively. Hence, the linear regression model confirms that different regions have varying effects on the listing price, with the USA having the most significant positive impact over Europe and Caribbean sailboats.
\subsection{Using Models to Evaluate Hong Kong Sailboats}
The Hong Kong (SAR) market presents a unique opportunity for the sailboat industry, given its strategic location, robust economy, and increasing interest in recreational boating. Therefore, in order to analysis the effect of our regression model to Hong Kong sailboats market, we first need to find the coefficient of Hong Kong region's listing prices. We then use the same web crawler, Octoparse, to collect $700$ more sailboats details specifically from the Hong Kong market, resulting a total of $4211$ data points. After that, we randomized these data points, repeat the same procedure in the previous section. But this time, our numerical variable representing the different region is: 1000 for Caribbean, 0100 for Europe, 0010 for USA, and 0001 for Hong Kong. After running the Linear Regression model, we successfully getting a linear regression model. The residual plot is shown below:
\begin{center}
    \includegraphics[scale = 0.315]{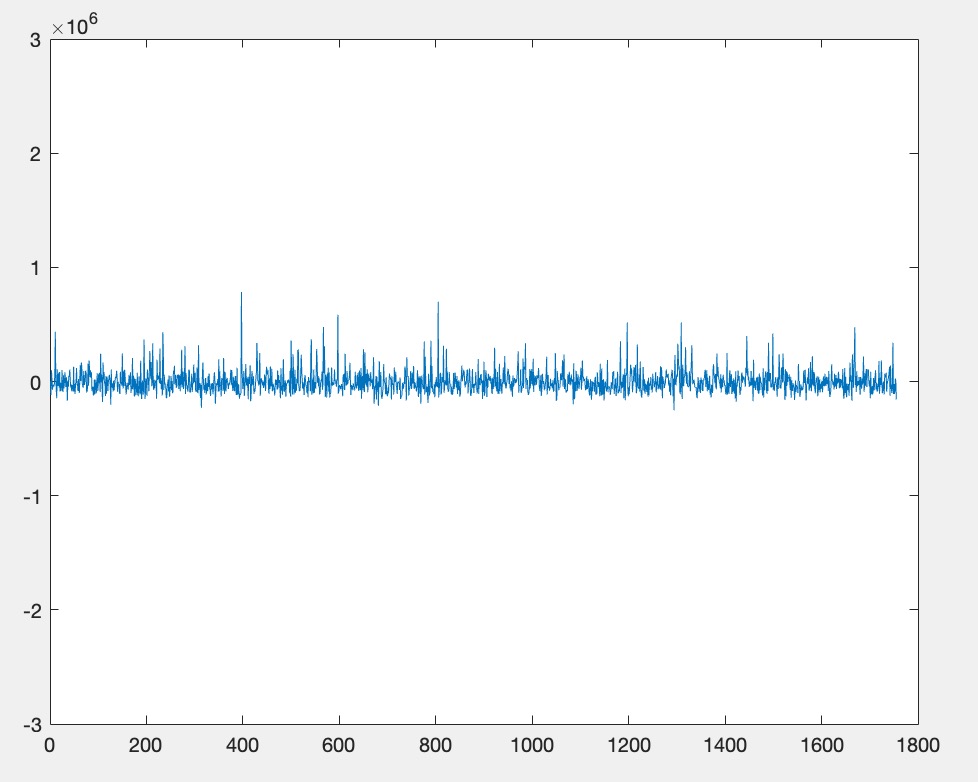} \\
    Figure 15
\end{center}
The present study aims to investigate the Hong Kong sailboat market using linear regression analysis. We observed that the residuals fall within the range of $-0.2 \cdot 10^6 - 0.3 \cdot 10^6$. The mean absolute error (MAE) obtained from MATLAB calculations is 53809, indicating acceptable accuracy for linear regression, although it was slightly higher prior to the inclusion of 700 data points from Hong Kong. Based on this result, we concluded that this model is suitable for analyzing the Hong Kong sailboat market. To conduct the analysis, a random sample of 3,000 sailboats from outside of Hong Kong, out of the total 4,210 sailboats, was selected and categorized into monohulls and catamarans. The sail price in Hong Kong was estimated by assigning a value of $1000$ to the numerical region variable for each boat in the sample. The sail prices of monohulls and catamarans were plotted separately and analyzed.
\begin{center}
    \includegraphics[scale = 0.45]{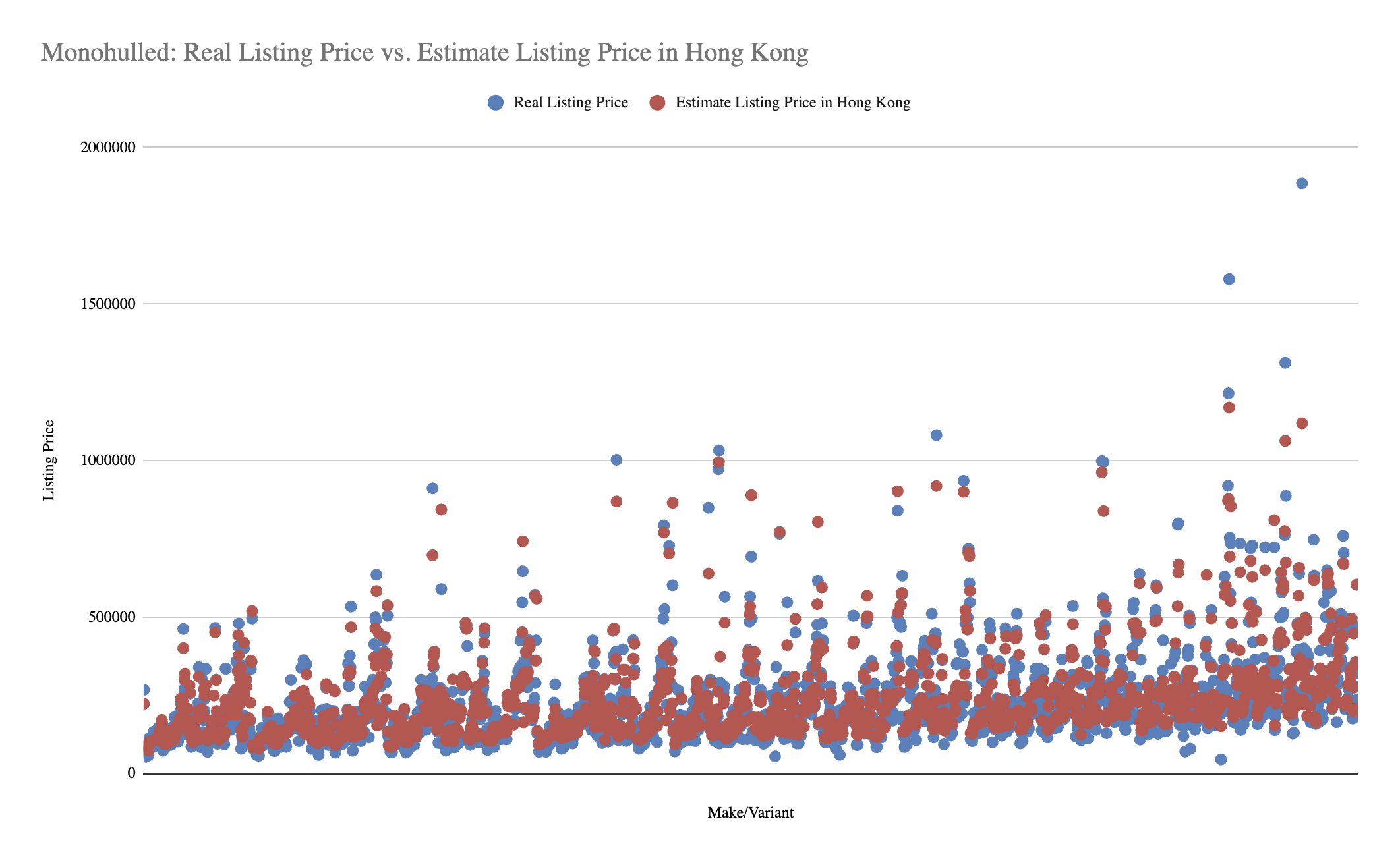}\\
    Figure 16 \\
    \includegraphics[scale = 0.45]{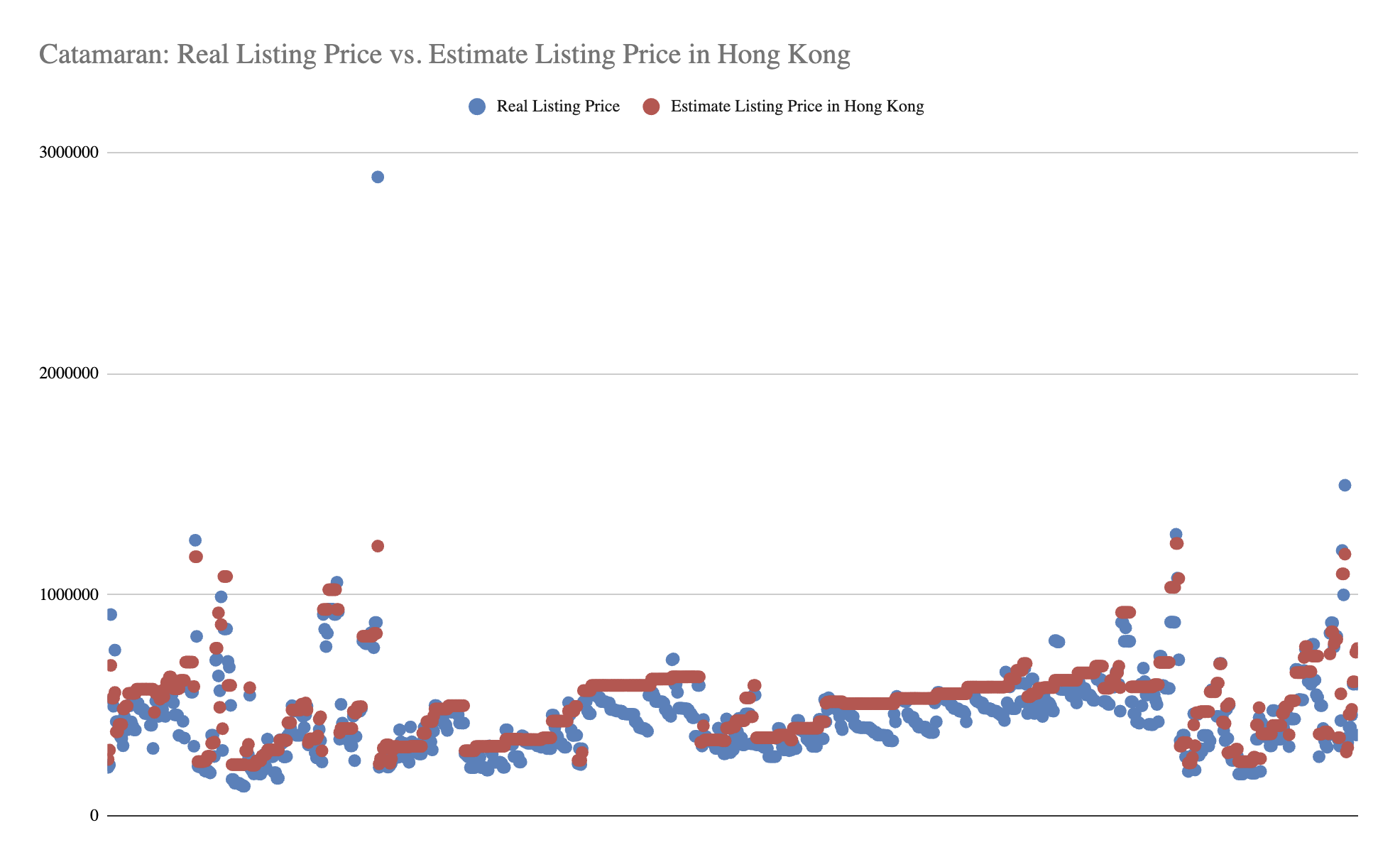}\\
    Figure 17\\
\end{center}
While the two comparison diagrams may not provide direct evidence of differences between the Hong Kong sailboat market and other global markets, our Linear Regression model reveals valuable insights. Specifically, the coefficient for the Hong Kong (1000) variable is estimated to be 16,804.39, suggesting that sailboat prices in Hong Kong are higher than those in the Caribbean, comparable to those in Europe, and lower than those in the USA. Therefore, this result provides compelling evidence that Hong Kong represents a unique and distinct market for sailboats. \\
\section{Validation and Sensitivity Analysis}
In accordance with Section 3, an evaluation of the sensitivity of our models was conducted through a 50 percent cross-validation procedure. Our previous study limited its assessment to the calculation of MAE and MSE on the testing group, but did not incorporate the testing group as a modeling group for validation purposes. To rectify this shortcoming, a comprehensive sensitivity analysis was performed on both models, which involved a complete cycle of testing that included the utilization of the testing group as a modeling group. \\
We first tested our gradient decent regression model used in section 3.1.5. As mentioned in that section, when testing the model on the testing group, the MSE and MAE are 13602449063 and 58032, respectively. This time, we re-implemented our model based on the other half of the group, and found out that the MSE and MAE on the other testing group is 14617295481 and 55973 respectively. Considering the high variance in our dataset, this level of float in MSE and MAE can be considered stabilized and valid. \\
Next, we conducted multiple variable linear regression analysis to investigate the impact of the region on the listing price. To evaluate the accuracy of our model, we employed a testing group as the modeling dataset. Our results indicate that the mean squared error (MSE) and mean absolute error (MAE) for the testing group were 15774541363 and 59128, respectively, where the corresponding values of 964296633 and 52209 was observed in the original analysis. Although the MSE and MAE values in our updated multiple variable linear regression model are slightly higher compared to the original analysis, the differences are not substantial enough to indicate a significant decline in the model's stability. In other words, the differences are within an acceptable range of error and do not substantially affect the model's overall performance or its ability to make accurate predictions. Therefore, we can still consider the model to be stable, and its outputs can still be used with reasonable confidence.\\
In conclusion, these findings from 50 percent cross validation highlights the robustness and accurate predictions of our two models.
\section{Interesting Features of The Dataset}
While building up models for problem one, we have preliminary discovered that the draft length of the sailboat has a negative correlation on the listing price. That means, a higher draft will result in a cheaper price of the sail boat. By definition, draft is how deep your boat's hull can go into the water. Our team hypothesized that a shallower draft may be more desirable for many buyers, as it allows for easier access to shallow water areas, such as coastal cruising, inland waterways, or anchoring close to the shore. This increased demand for boats with a shallower draft could drive up their prices. \\
As for the other technology specification in our dataset, all of which exhibited a normal positive correlation with the listing price. Specifically, we found that sailboat size, age, and hull type all had a significant impact on the final sale price. Notably, catamaran sailboats were consistently found to be more expensive than monohulled sailboats, despite similarities in technology specifications. Through a process of hypothesis testing, we propose that this price differential can be explained by the greater complexity of catamaran design, which requires more advanced technology, skilled labor, and expensive materials such as carbon fiber. Our findings suggest that the sailboat market is highly sensitive to a range of technical and design-related factors, and that these factors should be carefully considered by buyers and sellers alike.
\newpage
\section{Conclusion}
During this investigation, our research team initially employed a web-crawling algorithm, specifically Octoparse, to obtain essential technical specifications for the sailboats provided by COMAP, such as beam, draft, length of waterline, and sail area. Given that the original dataset comprised only 2,347 data points, an additional 1,331 sailboat data points were collected to supplement the analysis. The team implemented a gradient descent regression model to scrutinize the associations between predictor variables and sailboat prices, concurrently comparing it to multiple linear regression and ADADELTA models. The gradient descent regression model displayed the lowest mean squared error (MSE) and mean absolute error (MAE). A correlation analysis of sailboat types revealed that monohulled sailboats are generally more economical than catamarans.\\
To assess the connection between geographical region and sailboat pricing, multiple linear regression was utilized to determine the coefficients of the predictor variables. The three regions—Caribbean, Europe, and the United States—were incorporated as independent variables and categorized using index values stored in a categorical variable. This variable was subsequently converted into a virtual variable, facilitating its integration into the model to ascertain its relationship with pricing. Our findings suggest that Caribbean sailboats are generally less expensive than their European counterparts, which in turn are more affordable than those in the United States.\\
In a subsequent analysis, a web-crawler was employed to amass data for 700 sailboats listed for sale in Hong Kong. This new data was integrated into the original dataset, with the region categorical variable assigned a value of 4, subsequently converted to [0,0,0,1] as a virtual variable. A random selection of 2,000 sailboats not listed for sale in Hong Kong had their region variable altered to Hong Kong to evaluate the impact on the results. The comparative analysis revealed that, in relation to Hong Kong sailboat prices, those in the Caribbean were less expensive, while prices in Europe and the United States were marginally higher.

\pagebreak
\end{document}